%% file: main.tex
\documentclass[a4paper]{article}

\usepackage[english]{babel}
\usepackage[utf8x]{inputenc}
\usepackage[T1]{fontenc}
\usepackage{bbm}

\usepackage[a4paper,top=3cm,bottom=2cm,left=3cm,right=3cm,marginparwidth=1.75cm]{geometry}

\usepackage{amsmath}
\usepackage{graphicx}
\usepackage[colorinlistoftodos]{todonotes}
\usepackage[colorlinks=true, allcolors=blue]{hyperref}
\usepackage{xspace}
\usepackage{algorithm}
\usepackage{algpseudocode}
\usepackage{comment}
\usepackage{overpic}
\usepackage{epstopdf}
\usepackage{mathtools}
\usepackage{booktabs}
\usepackage{pgfplots}
\usepackage{courier}
\usepackage{textcomp}
\usepackage{array}

\makeatletter
\newcommand*{\ie}{{\em \@ifnextchar{.}{i.e}{i.e.\@~} }}
\newcommand*{\eg}{{\em \@ifnextchar{.}{e.g}{e.g.\@~} }}
\newcommand*{\etc}{{\em \@ifnextchar{.}{etc}{etc.\@~} }}
\newcommand*{\etal}{{\em \@ifnextchar{.}{et al}{et al.\@~}}}
\makeatother
\DeclareMathOperator*{\argmin}{arg\,min}

\newcommand{\net}{\mathit{f}} 
\newcommand{\disc}{\delta} 

\newcommand{\loss}{\mathit{l}}

\newcommand{\C}{\mathit{C}}
\newcommand{\TrainingIndexes}{\mathcal{T}}
\newcommand{\TrainingBatch}{T}
\newcommand{\UnseenIndexes}{\mathcal{U}}
\newcommand{\UnseenBatch}{U}
\newcommand{\gtLabel}{\lambda}
\newcommand{\gtLabelt}{\gtLabel_t}
\newcommand{\reffig}[1]{Figure~\ref{#1}}
\newcommand{\expect}[1]{\mathbbmtt{E}\left[#1\right]}
\title{An Adversarial Regularisation for Semi-Supervised Training of Structured Output Neural Networks}
\author{
\begin{tabular}[t]{>{\centering\arraybackslash}p{4cm}>{\centering\arraybackslash}p{4cm}>{\centering\arraybackslash}p{4cm}}
Mateusz Kozi\'nski & Lo\"ic Simon &Fr\'ed\'eric Jurie
\end{tabular}
{} \\
 \\
{\small Groupe de recherche en Informatique, Image, Automatique et Instrumentation de}\\
{\small Caen Normandie Univ, UNICAEN, ENSICAEN, CNRS, GREYC, 14000 Caen,
France} \\
 \\
 \begin{tabular}[t]{>{\centering\arraybackslash}p{4cm}>{\centering\arraybackslash}p{4cm}>{\centering\arraybackslash}p{4cm}}
 \small mateusz.kozinski@unicaen.fr  &\small  loic.simon@ensicaen.fr & \small frederic.jurie@unicaen.fr
 \end{tabular}
}
\date{}

\begin{document}
\maketitle

\begin{abstract}
\input{abstract.tex}
\end{abstract}


\input{intro.tex}
\input{related.tex}

\input{newmethod.tex}

\input{experiments.tex}

\input{conclu.tex}

\bibliographystyle{acm}
\bibliography{ref}

\end{document}

%% file: abstract.tex
We propose a method for semi-supervised training of structured-output neural networks. 
Inspired by the framework of Generative Adversarial Networks (GAN), we train a discriminator network to capture the notion of a `quality' of network output.
To this end, we leverage the qualitative difference between outputs obtained on the labelled training data and unannotated data. 
We then use the discriminator as a source of error signal for unlabelled data.
This effectively boosts the performance of a network on a held out test set.
Initial experiments in image segmentation demonstrate that the proposed framework enables achieving the same network performance as in a fully supervised scenario, while using two times less annotations.


%% file: intro.tex
\section{Introduction}

The unprecedented power that neural networks offer when applied to vision problems comes at a cost of large volumes of annotated training data. When the annotations are produced manually the process can be laborious and costly, especially for structured output problems like image segmentation. 

In this paper we propose an approach to semi-supervised training of structured output neural networks. The proposed approach allows to capitalize on large sets of unlabelled data. We show that the performance of a network trained in a fully supervised regime on a certain amount of labelled data can be matched by using a significantly smaller amount of labelled data in a semi-supervised setting, together with a sufficiently large volume of unlabelled data. In consequence, significant labelling effort can be saved.

In technical terms, we generate a useful error signal for data for which no ground truth labels are available, by means of adversarial training. During training, both the labelled training data and the unlabelled data is forwarded through the network. The network produces qualitatively better output on the labelled images than on the unlabelled images. Much like in training a Generative Adversarial Network (GAN), we train a discriminator network to capture this difference. The negative gradient of the discriminator with respect to its unlabelled input is used as the error signal for the unlabelled data. 

Our technical contribution consists in an adversarial learning approach for semi-supervised training of structured output neural networks.
A particular advantage of our method is that it can be applied to any structured output problem, independently of the architecture of the applied predictor.
Contrary to pre-training, the proposed method can be applied to a complete network, not just to its feature-extracting part.


%% file: related.tex
\section{Related work}
Our work is related to previous efforts to use unannotated data for training neural networks, including autoencoders, self-supervised learning and the use of GANs.

A considerable research effort has been devoted to autoencoders \cite{Hinton2006,Masci11} -  
neural networks that encode an image into a latent representation, from which they attempt to reconstruct the original image. Different regularization techniques are applied to impose useful properties on the hidden representation. The encoder of a trained autoencoder is considered a useful `feature extractor'. 
In a pre-training scenario \cite{Zeiler2011,Ranzato2007} the encoder of a trained autoencoder is incorporated as a feature extractor into another network, which is then fine-tuned for a particular task on labelled data. In a semi-supervised scenario \cite{Zhao2015,Zhang2016} parameters are shared between an encoder of an autoencoder and a feature extractor of a supervised network, and both are trained simultaneously. 
Plain autoencoders attempt to encode all the visual information in the latent representation. It has been hypothesised that much of the information is irrelevant for particular vision tasks, and autoencoders that transfer some of it between the input and the output \cite{Ranzato2007,Zeiler2011,Zhang2016}, instead of encoding everything in the latent representation, produce more useful representations. A recent example of such architecture is the ladder network \cite{Valpola2015,Rasmus2015}, where the critical information content that should be encoded in the latent representation is learnt in a semi-supervised setting.
One drawback of autoencoders is that they constrain the architecture of the supervised network to be the same as that of the encoder. While the convolutional autoencoders \cite{Masci11} with pooling layers \cite{Ranzato2007,Zeiler2011,Zhang2016} match the architectures of contemporary image classification networks \cite{vgg} well, they can only be matched to an initial part of a structured output network, for example one used for image segmentation \cite{conv_deconv,segnet}. In consequence, the other part of a network does not benefit from the unlabelled data. This is consistent with the intuition that the `final' part of such network, that upsamples feature maps, represents a correlation between the output variables. Such correlation cannot be learnt by an autoencoder that is never exposed to any ground truth annotations. 




A number of `self-supervised' methods of training feature extractors \cite{Doersch2015,Pathak2016,Noroozi2016} emerged recently. They consist in `deconstructing' unlabelled images by removing some information, and training a neural network to reproduce the original image. The `deconstruction' methods include masking image regions, or dividing an image into tiles and shuffling the resulting tiles. The corresponding reconstruction tasks include inpainting the masked regions based on the context \cite{Pathak2016} and guessing a relative position of two or more tiles \cite{Doersch2015,Noroozi2016}. The reconstruction requires extracting high-level information from the image, which makes the obtained `feature extractors' useful for other vision tasks. However, from the perspective of structured-output tasks, the `self-supervised' methods suffer from the same drawbacks as autoencoders: they constrain the architecture of the trained network and are not suitable for capturing dependencies between output variables.



Our unsupervised objective is inspired by the Generative Adversarial Networks (GANs) \cite{Goodfellow2014}. In GAN, a generator network is trained to transform a random vector originating from a simple sampling distribution to a sample from a complicated target distribution. The flagship application is to train a generator to yield realistically looking images from random vectors. The interesting property of GANs is that all that is required for training the generator is a collection of vectors originating from the target distribution. The error signal is backpropagated to the generator from a discriminator network that attempts to differentiate between vectors originating from the true target distribution and the ones generated by the trained network. The generator and the discriminator are trained in an alternating manner. Theoretically, GAN training has been shown to be an instance of a saddle point problem. GANs are difficult to train in practice, and some work has been devoted to improving their behaviour during training \cite{Salimans2016}, identifying architectures that work well in GANs \cite{Radford2015}, and generalizing the discriminator from a binary classifier to an energy function \cite{ebgan}.

A number of attempts at using GANs for unsupervised learning has been made recently \cite{Salimans2016,Radford2015}. In the simplest case, the initial layers of a discriminator are used as a feature extractor \cite{Radford2015}. In the task of image classification, the generated images can constitute a new class of input images \cite{Salimans2016}, augmenting the total number of training images.

GANs can also be used for mapping directly between two domains of interest. In this case discrimination is performed between pairs of input and output. A recent work \cite{pix2pix2016} showcased learning such a mapping between artistic depictions and photographs, images taken during day and night time, or segmentations and the corresponding images. The discriminator differentiates pairs of input and the corresponding ground truth output from pairs of input and output generated by the network. This new, learned cost function is shown to give more visually plausible results than the standard L2 reconstruction loss. However, it performed worse than the baseline loss on mappings with less output ambiguity, like the mapping from images to segmentation maps. The same type of loss has been demonstrated to boost segmentation results when combined with a standard cost function \cite{Luc2016}. The methods are fully supervised - the adversarial criterion is evaluated for the labelled training data. In contrast, we use a discriminator specifically to generate a training signal for unlabelled data.

Another body of research \cite{Donahue2016,ali} proves that mappings between the latent space and the data space can be learnt with just samples from both domains, and that corresponding input-output pairs are not necessarily needed. Two networks are trained simultaneously: a generator $g()$, producing samples from a latent representation $z$, and an encoder $f()$, inferring the latent representation from data $x$ coming from the target distribution. Discrimination is performed between two types of pairs: a pair of generated data and the corresponding latent vector $(g(z),z)$, and a pair of data originating from the target distribution, and the inferred latent representation $(x,f(x))$. At optimality $g()$ and $f()$ are proven to be inverses. However, the methods have only been shown to work on low-dimensional inputs and latent representations. 

The discrimination-based approach can also be applied to domain adaptation \cite{Ganin2015,hoffman2016}. The discriminator is learning to differentiate between features obtained for samples from two different domains, like synthetic and real images, or images representing different seasons of year. The discriminator is a source of an error signal, that makes the network produce similar features on data coming from both domains. 
The method can be used for object detection and image classification \cite{Ganin2015}, and for semantic segmentation \cite{hoffman2016}.
While the network architecture used in domain adaptation is similar to ours, the concept behind the methods is substantially different. The goal of domain adaptation is  to make the network insensitive to certain shift in the input data. In contrast, our goal is to regularize the network with use of the unlabelled data.

%% file: newmethod.tex
\section{Method description}
\begin{figure*}
\centering
\begin{overpic}[]{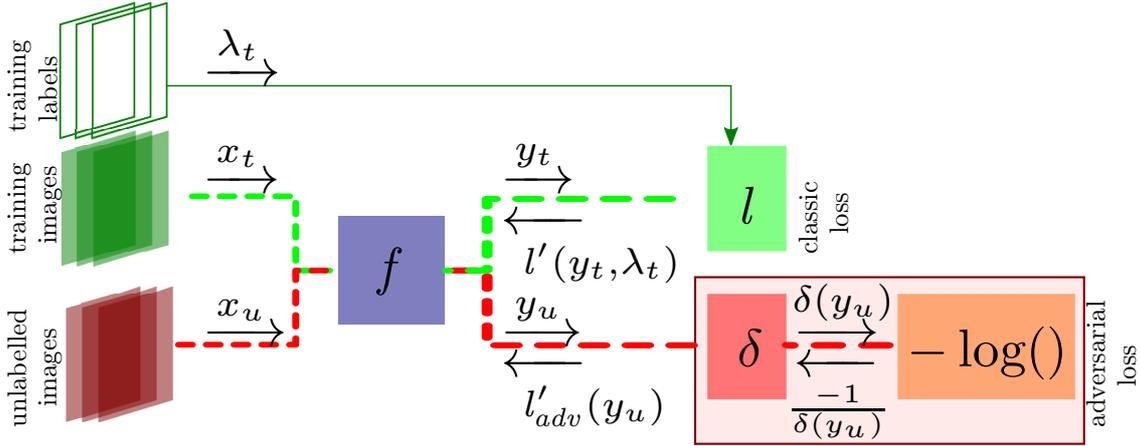}
\put( -5,  2){\rotatebox{90}{unlabelled}}
\put( -2,  4){\rotatebox{90}{images}}
\put( -5, 19){\rotatebox{90}{training}}
\put( -2, 20){\rotatebox{90}{images}}
\put( -5, 31){\rotatebox{90}{training}}
\put( -2, 32){\rotatebox{90}{labels}}
\put(72, 19){\rotatebox{90}{classic}}
\put(75, 21){\rotatebox{90}{loss}}
\put(100, 2){\rotatebox{90}{adversarial}}
\put(103, 6){\rotatebox{90}{loss}}
\put( 31, 16){\scalebox{2}{$\net$}}
\put( 66, 22){\scalebox{2}{$\loss$}}
\put( 82.5,8){\scalebox{2}{$-\log()$}}
\put( 66,  8){\scalebox{2}{$\disc$}}
\put( 14, 25){\scalebox{2}{$\xrightarrow{{x_{t}}}$}}
\put( 14, 10){\scalebox{2}{$\xrightarrow{{x_{u}}}$}}
\put( 14, 35.5){\scalebox{2}{$\xrightarrow{\gtLabelt}$}}
\put( 43, 25){\scalebox{2}{$\xrightarrow{{y_{t}}}$}}
\put( 43, 10){\scalebox{2}{$\xrightarrow{{y_{u}}}$}}
\put( 43, 21){\scalebox{2}{$\xleftarrow[{\mathmakebox[0pt][l]{\loss'(y_t,\gtLabelt)}}\phantom{\C'}]{}$}}
\put( 43,  7){\scalebox{2}{$\xleftarrow[{\mathmakebox[0pt][l]{\loss'_\mathit{adv}(y_u)}}\phantom{\C'}]{}$}}
\put( 71.5, 10){\scalebox{2}{$\xrightarrow{\mathmakebox[0pt][l]{\!\!\disc(y_{u})}\phantom{\C'}}$}}
\put( 71.5,  7){\scalebox{2}{$\xleftarrow[{\mathmakebox[0pt][l]{\!\!\!\!\frac{-1}{\disc(y_u)}}}\phantom{\C'}]{}$}}
\end{overpic}
  \label{fig:architecture}
\caption{The flow of data and error signals when training a structured output network $\net$ with the proposed method, presented in algorithm \ref{ourBasicTrainingAlg}. The discriminator update is not shown in the drawing. The green line denotes the flow of labelled training data and the corresponding gradients. The red line denotes the flow of unlabelled data and the corresponding gradients. By $\loss'(y,\gtLabel)$ we denote the partial derivative of the loss with respect to the prediction, $\loss'(y,\gtLabel)=\nabla_{y} \loss(y,\gtLabel)$.}
\end{figure*}

We address the problem of training a structured output network $\net_w$, parametrised with a weight vector $w$, to produce predictions $y$ on input data $x$.
The $x$ and $y$ can take any form as long as they can be input and output by a neural network.
We are targeting a scenario in which, in addition to a set of training examples $x_t$, $t\in\TrainingIndexes$, with annotations $\lambda_t$, a volume of unlabelled examples $x_u$, for $u\in\UnseenIndexes$, is available. 
To handle the unlabelled data, our approach combines a classic supervised loss $\loss(y_t,\lambda_t)$, measuring the consistency of $y_t$ and $\lambda_t$, with a novel and unsupervised one $\loss_\mathit{adv}(y_u)$. The new loss term is described in section \ref{sec:loss}. We define a global cost consisting of two components
\begin{equation} \label{fullCost}
  \begin{split}
    \C_\mathit{tot}(w) =& \C(w) + \alpha\C_\mathit{adv}(w) \\
    =& \expect{\loss(f(x_t),\lambda_t)} + \alpha \expect{\loss_\mathit{adv}(f(x_u)},
  \end{split}
\end{equation}
where $\alpha$ is a constant.
Training consists in determining the optimal network parameter by solving
\begin{equation} \label{fullObjective}
w^* = \argmin_{w} \C_\mathit{tot}(w) .
\end{equation}
We describe the training algorithm in section \ref{sec:alg}



\subsection{Adversarial loss \label{sec:loss}}
Defining a loss function, that measures network performance based only on its outputs seems infeasible. On the other hand, given two image segmentations, output by a neural network at two sufficiently distant training epochs, a human can spot the perceptual difference corresponding to increasing accuracy. This suggests that there might exist a measure of `output quality', applicable at least for a certain range of accuracy. In this section, we attempt to construct such a function.

In a classical setting, training a network on the labelled training data $(x_t,\lambda_t)$, $t\in\TrainingIndexes$ results in a qualitative difference between outputs $y_t=\net_w(x_t)$ and outputs produced for the unseen data $y_u=\net_w(x_u)$, $u \in\UnseenIndexes$. Ideally, $x_t$ and $x_u$ are identically distributed, so one might think that the same holds for $\net_w(x_t)$ and $\net_w(x_u)$.
In practice, the dimensionality of $x$ is typically large and the training set is not necessarily representative of the variability present in some unseen data. This biases $\net_w$ towards better performance on the training examples.\footnote{Note that we are not referring to the phenomenon of overfitting, where a mismatch of model complexity to the size of the training set can cause a situation where decreasing the training objective results in increasing the error on a held out test data, but simply to the higher performance of the network on the training set.} We leverage this qualitative difference to define the unsupervised cost $\C_\mathit{adv}$ as a regularisation term that tends to close this gap.

Inspired by the GAN framework, we propose to train a discriminator network $\disc_v$, parametrised by $v$, to capture the qualitative difference between $\net_w(x_t)$ for $t\in\TrainingIndexes$, and $\net_w(x_u)$, for $u\in\UnseenIndexes$.
We use a binary discriminator. We interpret the scalar output $\disc_v(y)$ as the likelihood that $y$ has been obtained from an element of the labelled training set $y=\net_w(x_t)$, $t\in\TrainingIndexes$, and we interpret $1-\disc_v(y)$ as the likelihood of $y$ originating from the unlabelled set $y=\net_w(x_u)$, $u\in\UnseenIndexes$. The optimal parameter configuration of the discriminator $v^* = \argmin_{v}CE_\mathit{disc}(v)$ is defined in terms of the cross-entropy $CE_\mathit{disc}(v)=\left(-\expect{\log(\disc_v(f(x_t)))} - \expect{\log(1-\disc_v(f(x_u)))}\right)$.

When the discriminator is trained to optimum, its negative logarithm can be used as a local `quality measure' for image segmentations. This is because $\disc_{v^*}(y)$ is the likelihood that $y$ has been generated on the training set, and the outputs on the training set are qualitatively better. We therefore define the unsupervised cost as
\begin{equation} \label{advCost}
  \begin{split}
    C_\mathit{adv}(w)=&\expect{-\log(\disc_{v^*}(f_w(x_u))}.
  \end{split}
\end{equation}
Minimising \eqref{advCost} with respect to $w$ drives $\net_w$ towards reducing the gap between performance on labelled training and unlabelled data. It is however important to do so in a one-way manner, so as to avoid making the performance of $\net_w$ degrade on the labelled examples. Clearly, we want $\net_w$ to perform as well on unlabelled examples as it does on labelled ones but not the other way around. Therefore, we apply the adversarial component of the cost function only to the unlabelled data.
%

\subsection{Algorithm \label{sec:alg}}
The minimization can be performed with a gradient-based optimization routine, for example SGD. The gradient of the objective consists of two components and its estimate on a training batch $\TrainingBatch$ and unlabelled batch $\UnseenBatch$ can be denoted as
\begin{equation} \label{fullGradient}
\nabla_w \C^{\TrainingBatch\UnseenBatch}_\mathit{tot}(w) = \nabla_w \C^{\TrainingBatch}(w) + \alpha \nabla_w \C^{\UnseenBatch}_\mathit{adv}(w) .
\end{equation}
The gradient can be computed by backpropagation. 
The flow of data and gradients forward and back through the networks is depicted in \reffig{fig:architecture}.
In practice, we train the network using algorithm \ref{ourBasicTrainingAlg}. The $\mathrm{update}(w,g)$ procedure accepts the network weights $w$ and a gradient of the cost function $g$ and performs an update on $w$. While we used SGD with momentum, any update rule used for training neural networks is applicable. Instead of training the discriminator to optimality at each iteration, we perform $k$ updates of the discriminator for a single update of the network $\net_w$ itself. There is no guarantee of convergence of the algorithm. However, our experiments demonstrate its practical utility. 

\begin{algorithm} 
\caption{Training a structured output network with adversarial cost for unlabelled data \label{ourBasicTrainingAlg}}
\begin{algorithmic}[1]
  \State{$v,w\leftarrow \mathrm{randInit}()$}
\While{not converged}
  \For{$\mathit{IterNum}=1$ to $k$} 
    \State{$T\leftarrow \mathrm{pickBatch}\big(\TrainingIndexes, batchSize\big)$}
    \State{$U\leftarrow \mathrm{pickBatch}\big(\UnseenIndexes, batchSize\big)$}
    \State{$g\leftarrow -\sum_{t\in T}\nabla_v \log(\disc_v(\net_w(x_t))) - \sum_{u\in U}\nabla_v \log(1-\disc_v(\net_w(x_u)))$}
  \State{$v\leftarrow \mathrm{update}\big(v, g\big)$ } 
  \EndFor 
  \State{$T\leftarrow \mathrm{pickBatch}\big(\TrainingIndexes, batchSize\big)$}
  \State{$g_T\leftarrow \sum_{t\in T}\nabla_w \loss(\net_w(x_t))$}
  \State{$U\leftarrow \mathrm{pickBatch}\big(\UnseenIndexes, batchSize\big)$}
  \State{$g_U\leftarrow - \sum_{u\in U}\nabla_w \log(\disc_v(\net_w(x_u)))$}
  \State{$w\leftarrow \mathrm{update}\big(w,  g_T+\alpha g_U\big)$}
\EndWhile
\end{algorithmic}
\end{algorithm}

%% file: experiments.tex
\section{Experimental evaluation}

The goal of the experiments is to compare semi-supervised training with the adversarial loss to fully supervised training. In particular, we are interested in the trade-off between the labelling effort and the performance of the trained network. The question we are asking is: given a collection of training input data and having labelled a part of it, what is the benefit, in terms of network performance, of labelling a certain portion of it with respect to using it in an unsupervised manner. Knowing the answer helps to take the decision whether the expense of manpower required to label a certain portion of data is worth the increase in performance that it can bring.

We run experiments according to the following pattern. We run the baseline method on the whole training set, and on the training set consisting of $\frac{1}{2}$, $\frac{1}{4}$ and $\frac{1}{8}$ images of the original training set. Then, we apply the proposed method, with $\frac{1}{2}$, $\frac{1}{4}$ and $\frac{1}{8}$ of the training set used for the supervised subtask, and the remaining part of the training set used in an unsupervised manner.

\begin{table*}
\centering
\caption{Performance of segnet-basic on CamVid w.r.t.~the fraction of used annotations. The baseline is a fully supervised SGD with momentum. The proposed method uses the reported fraction of annotated data and the remaining part of the training set without annotations. The performance measures: intersection over union (IoU), average per-class recall (C) and global precision (G).\label{resultsExp1}}
\setlength{\tabcolsep}{3pt}
\centering
\begin{tabular}{l ccc c ccc c ccc c ccc}
\rule{0pt}{6ex}
subs. factor:& \multicolumn{3}{c}{1} && \multicolumn{3}{c}{$\frac{1}{2}$} && \multicolumn{3}{c}{$\frac{1}{4}$} && \multicolumn{3}{c}{$\frac{1}{8}$} \\
         & IoU  & C         & G        && IoU   & C   & G  && IoU    &  C   & G   && IoU  & C    & G    \\
baseline & 49.3 &      64.6 &      83.5 && 46.2 & 58.0 & 82.6&& 44.3 & 56.7 & 81.0&& 40.0 & 50.9 & 79.1 \\
ours&\textemdash&\textemdash&\textemdash&& 47.9 & 60.0 & 83.4&& 47.3 & 58.8 & 82.3&& 43.0 & 53.3 & 81.9 \\
\end{tabular}
\end{table*}
We use the CamVid dataset, in the version used by Badrinarayanan, Kendall and Cipolla \cite{segnet}. It consists of images captured by a forward-looking vehicle-mounted camera, of the size of $360\times 480$ pixels. The datasets contains of 367 training, 101 validationframe and 233 test images. The set of labels consists of 11 classes. The compact size of the dataset lets us run a number of experiments in reasonable time.

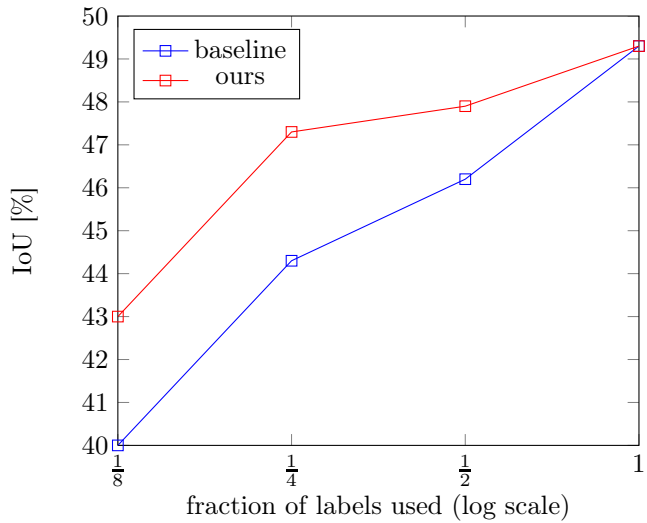
\begin{figure}
\centering
\begin{tikzpicture}
\begin{semilogxaxis}[
    xlabel={fraction of labels used (log scale)},
    ylabel={IoU [\%]},
    xmin=0.125, xmax=1,
    ymin=40, ymax=50,
    xticklabels={$\frac{1}{8}$,$\frac{1}{4}$,$\frac{1}{2}$,$1$},
    xtick={.125,.25,.5,1},
    ytick={40,41,42,43,44,45,46,47,48,49,50},
    legend pos=north west,
    ymajorgrids=false,
    grid style=dashed,
]
\addplot[
    color=blue,
    mark=square,
    ]
    coordinates {
    (.125,40)(.25,44.3)(.5,46.2)(1,49.3)
    };
    \addlegendentry{baseline}
\addplot[
    color=red,
    mark=square,
    ]
    coordinates {
    (.125,43)(.25,47.3)(.5,47.9)(1,49.3)
    };
    \addlegendentry{ours}
\end{semilogxaxis}
\end{tikzpicture}
\caption{The IoU attained by segnet-basic on the CamVid dataset with respect to the number of annotations used for training. Note that the proposed method almost matches the baseline with two times as many annotations. \label{iouplot1}}
\end{figure}
We use the segnet-basic network \cite{segnet}. It has an encoder-decoder architecture, where the encoder consists of four blocks of architecture \texttt{64c(7)-BN-ReLU-MP(2)}, where \texttt{Nc(K)} denotes a layer of \texttt{N} convolutional filters of size \texttt{K}\texttimes\texttt{K}, applied with output stride of one pixel in both directions, \texttt{BN} denotes the batch normalization, \texttt{ReLU} denotes the Rectified Linear Unit and \texttt{MP(K)} denotes the max-pooling operation performed in windows of size \texttt{K}\texttimes\texttt{K} with the output stride of \texttt{K} pixels. Each encoder block effectively subsamples the feature map by a factor of 2 in both dimensions. The decoder consists of four blocks of \texttt{MU(2)-64c(7)-BN}, where \texttt{MU(K)} denotes a max-unpooling layer, with the output stride of \texttt{K}\texttimes\texttt{K}, where the unpooling indeces are transfered from the symmetric max-pooling layer in the encoder. Each block of the decoder effectively upsamples the feature map by a factor of 2 in both dimensions. The output is produced by a 1x1 convolutional layer with 11 filters and a stride of 1 in both directions. We refer the reader to the original work \cite{segnet} for a more detailed explanation of the encoder-decoder architecture with coupled pooling and unpooling layers.

The discriminator consists of three blocks of \texttt{64c(3,2)-BN-LReLU}, followed by a global average pooling layer and the final linear layer with a single output variable. By \texttt{64c(3,2)} we denote a convolutional layer consisting of 64 filters of size \texttt{3}\texttimes\texttt{3} and an output stride of  \texttt{2}\texttimes\texttt{2}, and \texttt{LReLU} denotes a rectified linear unit with the slope of $0.2$ for the `deactivated' piece.
When performing the experiments we found out that an important aspect of the discriminator architecture is the global pooling layer. Its effect is similar to per-pixel discriminator ground truth used in \cite{pix2pix2016} and consists in preventing the discriminator from overfitting by memorizing the contents of labelled training and unlabelled images. 
The discriminator is binary and trained using a cross-entropy loss.

We use the basic SGD algorithm for updating both the segmentation network and the discriminator. We use momentum of $0.9$ and weight decay of $0.001$. For both the baseline and the proposed approach we train the network for $10^4$ iterations with a learning rate of $0.1$, then for $4 \cdot 10^3$ iterations with a learning rate of $0.05$, another $4 \cdot 10^3$ with a learning rate of $0.025$ and finally for $2 \cdot 10^3$ iterations with a learning rate of $0.0125$. We jitter the training images by shifting them by a random number of between $0$ and $32$ pixels in both directions. We perform the accuracy tests on the network defined by the weight vector after the last update in this procedure, instead of cherry-picking the model using a cross validation on the validation set. We found out that this strategy gave better results for both the baseline and the proposed algorithm.

When using our method, we set $k=1$, that is, we update the discriminator once per every update of the trained network. We use batches of $16$ training images for the baseline, and batches of $8$ training and $8$ unlabelled images for the semi-supervised setting.

We present numerical results in table \ref{resultsExp1} and in figure \ref{iouplot1}. The baseline attains an accuracy of $49.3\%$ Intersection-over-Union (IoU), which exceeds the performance of $47.7\%$ reported in the original paper \cite{segnet}. We suspect the increase comes from the differences in the training protocol, including jitter and a decreasing learning rate. Our method consistently outperforms the baseline. Besides 
for every ratio of supervision, the performance 
of our network is nearly as good as the baseline using twice as many labelled examples. For a ratio of $\frac 14$, our approach even improves on the baseline with $\frac 12$ annotations used.

\begin{table}
\centering
\caption{The impact of the weight decay factor on the accuracy of segnet-basic trained on $\frac{1}{8}$-th of the full CamVid training set, compared to the result of the proposed, adversarial regularisation\label{resultsRegularisation}}
\setlength{\tabcolsep}{3pt}
\centering
\begin{tabular}{l cccccc c c}
\rule{0pt}{6ex}
             & \multicolumn{6}{c}{weight decay alone}     & & ours \\
decay factor & 0    & 5e-4 & 1e-3 & 5e-3 & 1e-2 & 5e-2 & & 1e-3\\
IoU          & 38.5 & 38.5 & 40.0 & 40.0 & 39.5 & 29.8 & & 43.0
\end{tabular}
\end{table}
\paragraph{Comparison to weight decay} Since the proposed adversarial loss can be interpreted as a regularization, it makes sense to ask how it compares to standard regularization techniques. While we already used weight decay in all our experiments, it can be argued that decreasing the training set should be accompanied by increasing the regularisation factor. We run an experiment to investigate this claim. As in the previous experiments we use weight decay, which is equivalent to adding a factor of $\frac{\beta}{2}|w|_2$ to the objective, where $\beta$ is the weight decay factor. We run several experiments with different values of $\beta$ according to the protocol used in the previous experiment. We only run training on the smallest training set, consisting of $\frac{1}{8}$ images of the original training set, since we expect the regularisation to play larger role for smaller training sets. We present the results in table \ref{resultsRegularisation}. We observe that weight decay has impact on improving the test performance - setting $\beta=1e-3$ increases the IoU by $1.5$ percent points with respect to the IoU attained by a network trained without regularisation. However, this seems to be the limit of the positive influence of weight decay. Adding our semi-supervised component to the objective, already regularised with weight decay, boosts the performance by further $3$ percent points.

%% file: conclu.tex
\section{Conclusion}
\label{sec:conclusion}

In this work, we propose a novel kind of regularization technique that can be applied in a semi-supervised context. Our approach attempts to learn a good
regularization cost that is guided by an adversarial scheme. The rational behind this 
choice is that, as a standard learning procedure goes, a trained network tends to perform better on training than on test data. We can
therefore train a classifier to discriminate between labelled and unlabelled output
distributions. Then, the likelihood estimate produced by the discriminator can be
used as a signal to improve the behaviour of the main network on the unlabelled population.

We have leveraged the aforementioned principle to derive a generic framework that 
can be applied to a large number of typical machine learning problems where 
a structured output is generated. In order not to hinder the exposition, we have 
focused our experiments on a single study case, namely semantic segmentation. 
Nonetheless, our approach can be adapted seamlessly to different tasks,
such as depth or normal inference. In the considered scenario,
we have evaluated our method on a standard benchmark and demonstrated that
our regularization achieves substantial improvements over a baseline using
only labelled data. We have also studied the evolution of the performance 
in response to a varying ratio of supervision.
One interesting observation of this 
study is that, when sufficient number of unlabelled examples is available, 
the amount of annotations required to attain certain performance is two times smaller for our semi-supervised approach than for the fully-supervised baseline.